\documentclass[runningheads]{llncs}
\usepackage[T1]{fontenc}
\usepackage{amsmath,amssymb,amsfonts}
\usepackage{dsfont}
\usepackage{algorithmic}
\usepackage{graphicx}
\usepackage{nicefrac}       
\usepackage{microtype}      
\usepackage{textcomp}
\usepackage{url}
\usepackage{float}
\usepackage[dvipsnames]{xcolor}
\usepackage{tabularx}
    \newcolumntype{L}{>{\raggedright\arraybackslash}X}
\usepackage{multirow}
\usepackage{adjustbox}
\usepackage{diagbox}
\usepackage{comment}
\usepackage{cite}
\usepackage{tikz}
\usepackage{hyperref}
\usetikzlibrary{patterns}
\usetikzlibrary{fit,positioning}
\usetikzlibrary{shapes.geometric, calc}
\usepackage{fourier} 
\usepackage{array}
\usepackage{subcaption} 
\usepackage{makecell}
\usepackage{tabstackengine}
\usepackage{dblfloatfix}
\usepackage{booktabs}
\usepackage{wrapfig}
\usepackage{orcidlink}
\def\BibTeX{{\rm B\kern-.05em{\sc i\kern-.025em b}\kern-.08em
    T\kern-.1667em\lower.7ex\hbox{E}\kern-.125emX}}

\definecolor{mypink2}{RGB}{219, 48, 122}
\definecolor{cornflowerblue}{rgb}{0.39, 0.58, 0.93}
\definecolor{dollarbill}{rgb}{0.52, 0.73, 0.4}
\begin{document}
\title{Study Design and Demystification of Physics Informed Neural Networks for Power Flow Simulation}
\titlerunning{Demystifying Physics Infromed Neural Networks for Power Flow}
%
\author{Milad Leyli-abadi\inst{1}\orcidlink{0000-0001-8006-5522} \and
Antoine Marot\inst{2}\orcidlink{0000-0003-1753-5493} \and
Jérôme Picault\inst{2}\orcidlink{0009-0009-0727-3048}}


\institute{IRT SystemX, Palaiseau, France\and
RTE, Paris, France
\\
\email{milad.leyli-abadi@irt-systemx.fr}
}
\maketitle              
\begin{abstract}
In the context of the energy transition, with increasing integration of renewable sources and cross-border electricity exchanges, power grids are encountering greater uncertainty and operational risk. Maintaining grid stability under varying conditions is a complex task, and power flow simulators are commonly used to support operators by evaluating potential actions before implementation. However, traditional physical solvers, while accurate, are often too slow for near real-time use. Machine learning models have emerged as fast surrogates, and to improve their adherence to physical laws (e.g., Kirchhoff’s laws), they are often trained with embedded constraints—known also as physics-informed or hybrid models. This paper presents an ablation study to demystify hybridization strategies, ranging from incorporating physical constraints as regularization terms or unsupervised losses, and exploring model architectures from simple multilayer perceptrons to advanced graph-based networks enabling the direct optimization of physics equations. Using our custom benchmarking pipeline for hybrid models called LIPS, we evaluate these models across four dimensions: accuracy, physical compliance, industrial readiness, and out-of-distribution generalization. The results highlight how integrating physical knowledge impacts performance across these criteria. All the implementations are reproducible and provided in the corresponding Github page\footnote{\url{https://github.com/Mleyliabadi/pinns-powergrid}}.

\end{abstract}
\section{Introduction}
To enable the power grid management and ensure the stability, the grid operators should continuously monitor the infrastructures and suggest remedial actions when necessary. However, 
such actions cannot be applied directly on the grid without any prior impact analysis as they
may cause damages in the worst-case scenarios. The integration of renewable energy sources, e.g., wind and solar, further increased the uncertainty level and required the power systems to be flexible \cite{babatunde2020power}. Hence, power flow simulations are extensively used to help grid operators in their decision-making process. The simulations allows to study different scenarios or solutions  and simulate the impact of their actions in a digital environment, before their application on the real grid. 

In the real-world scenarios, thousands of these simulations may be required, to avoid any nonlinear effect of some local changes. The power flow simulations are currently based on the resolution of the physical equations which are generally solved using iterative numerical optimization methods such as Newton-Raphson \cite{stott1974review}. Although these methods have the advantage of being precise and compliant with respect to physical constraints, their exponential time complexity does not allow their use in a large-scaled grid.

In this context, various studies have been interested in the use of machine learning-based approaches \cite{carleo2019machine}. They have also been used for power flow computation \cite{hasan2020survey, donnot2019leap}. The graph neural networks are also successfully used for this purpose \cite{battaglia2018relational, lin2024powerflownet}. Their fast inference time and leveraging the properties derived from graph theories may be a key-enabler to replace the complex physics-based solvers. Among the disadvantages of these approaches, we may cite their generalization problems (e.g., grid scale-up and unseen data), physics noncompliance and to be dependent on the quality and the volume of the data.

Recently, the physics-informed neural networks (PINNs) \cite{raissi2017physics} are successfully applied in various domains to solve physics-based problems such as fluid dynamics \cite{tompson2016}, scientific simulations \cite{kasim2021building} and weather forecasting \cite{lam2023learning}. Various similar works are also conducted for power flow simulations \cite{donon2020neural,lopez2023power, lei2020data}. As such, the physical constraints may be leveraged in different ways to make the data-driven approaches more compliant to the physical context. Among the most used strategies we can cite the direct use of the physics equations as the optimization objective during the training \cite{donon2020neural}, warm start points using machine learning followed by optimization \cite{xavier2021learning, baker2019learning}, or their use as a regularization term. However, each of these strategies comes with their pros and cons. Most of the studies use one or another, but do not report and compare their results using a set of standardized and comprehensive evaluation criteria.


In this paper, we try to bridge this gap; that is, to design a set of step-by-step studies to show the impacts of considering physical constraints whether as a regularization term in a simple neural network (MLP), or by using a two-stage approach where the warm start points inferred by an MLP are fed to a message passing graph-based approach optimizing directly the physics equations.  
We also suggest evaluating performance using an extensive benchmark through various evaluation criteria categories based on our Learning Industrial Physical Simulation (LIPS) framework \cite{leyli2022lips}. In addition to traditional machine learning metrics, using this standardized framework, we evaluate the performance with respect to physics compliance, out-of-distribution generalization and industrial readiness. All the studies provided in this paper are also available on the Github repository. 
All the datasets used for the training and evaluation of models are designed and generated for the sake of this study by considering real-world scenarios and could easily be reproduced using the provided scripts.

In different designed studies, we consider the steady-state hypothesis of observations by considering various grid configurations. These configurations may reflect the actions that the grid operators may take in different scenarios to keep the grids in a stable state. As this works aims to be also didactic and simple, we use a Direct Current (DC) solver for the generation of different scenarios.
We make use also of two different grid sizes for the experiments as a way to evaluate the scale-up capability of different benchmarked approaches.

The rest of the paper is organized as follows. Section \ref{sec:problem} describes the general framework of the study including the power flow simulation problem and the technical details of the generated datasets. Section \ref{sec:study} provides the details of different studies, along with graphical visualizations, that are designed to solve the power flow problem whether using pure machine learning techniques or their physics-informed variants. Section \ref{sec:benchmark} presents the experimental settings and the corresponding benchmark results. Finally, Section \ref{sec:conclusion} concludes the paper by highlighting insights for future works.

\section{Problem framework and datasets}
\label{sec:problem}
As mentioned, to operate the power grid in near real-time, a high number of power flow simulations are required by the operators to study the impact of the potential remedial actions. In addition to the industrial requirements such as fast simulations and scale-up capabilities, these simulations should be representative of real-world power grid dynamics, i.e., they should be reliable in various configurations where the data distribution could be slightly changed. For the models to be compliant to physical laws, they should consider them as constraints during the optimization. The next section presents the technical details of physical solvers which are the mostly used tools for simulations of power flows. It follows by an introduction of the datasets that are generated for the purpose of this study. Table \ref{tbl: notation} presents the notations that are used throughout the paper.

\begin{table}[htb]
\caption{Notation table}
\scalebox{1}{
\begin{tabular}{l|l}
Symbol & Description \\ \hline
$i$ & \text{Index associated with an observation (sample)} \\
$k$ & \text{Index associated with a grid node}\\
$\ell$ & \text{Index associated with a power line}\\
$or$ & \text{Index associated with the "origin" extremity of a power line by convention}\\
$ex$ & \text{Index associated with the second extremity of a power line by convention}\\
$p_k$ & \text{Active power at node $k$}\\
$q_k$ & \text{Reactive power at node $k$}\\
$v_k$ & \text{Voltage at node $k$}\\
$\theta_k$ & \text{Voltage phase (phasor) at node $k$}\\
$p^\ell$ & \text{Active power flow over line $l$}\\
$q^\ell$ & \text{Reactive power flow over line $l$}\\
$a^\ell$ & \text{Current over line $l$}\\
$\tau$ & \text{Topology vector indicating the bus connectivity}\\
$L$ & \text{Number of power lines in the grid}\\
$K$ & \text{Number of substations (nodes) in the grid}\\
$B$ & \text{Number of buses in the grid}\\
$N$ & \text{Number of samples in training set} \\
$M$ & \text{Number of samples in test set} \\
\end{tabular}}
\label{tbl: notation}
\end{table}

\subsection{Physical solvers}
Physical solvers which are currently used for the grid simulations are based on the resolution of power flow equations \cite{vlach1983computer, saadat1999power}. These equations for the alternative current (AC) are given by:
\begin{eqnarray}
\label{eq:Power_equations}
\left\{
\begin{array}{lll}
0 = -&{p_k} + \sum_{m=1}^K {|v_k|} {|v_m|} (g_{k,m}\cdot \cos ({\theta_k} - {\theta_m}) + b_{k,m} \sin ({\theta_k} - {\theta_m})) & \text{Active power;}\vspace{0.2cm} \\
0 = &{q_k} + \sum_{m=1}^K {|v_k|} {|v_m|} (g_{k,s}\cdot \sin ({\theta_k} - {\theta_m}) - b_{k,m} \cos ({\theta_k} - {\theta_m})) & \text{Reactive power,}
\end{array}
\right.
\end{eqnarray}
where $g_{km}$ and $b_{km}$ are real and imaginary parts of the admittances connecting the nodes $k$ and $m$. To solve these equations, an iterative optimization algorithms like Newton-Raphson may be used. Such iterative methods require a high number of steps to be converged and do not satisfy the near real-time requirement of power flow simulations. Hence, several contributions in state-of-the-art are interested in the use of physics-informed machine learning to cope with this problem. However, none of these papers provides an in-depth analysis of considering the physical knowledge when using machine learning algorithms based on neural networks.

To simplify the power flow simulation problem, in this paper we suggest to use the direct current (DC) power flow system which reduces the physical equations to:
\begin{equation}
    \label{eq:ac_powerflow}
    0 =\ \  -{p_k}\ \  - \sum_{\substack {j=1,}\\
                       j\neq k}^K b_{kj}(\theta_k - \theta_j).
\end{equation}
The DC simplifies the AC equations by assuming small phasors (voltage angles) differences and implying constant bus voltages (1 p.u.). DC does not generate reactive power. This means that, in a DC system, only real (or active) power is transmitted. These make the DC solver significantly faster to compute in comparison to AC solvers.

\subsection{Datasets}
Two different power grids with different sizes are investigated in this paper. Their characteristics are presented in Table \ref{tab:envs}. The first ones includes a set of 14 nodes and 20 power lines. Due to its small size, the non-linear impact of topological changes propagates more severely throughout the grid. To gain insight into the scalability of the models, a slightly larger grid with 36 nodes and 59 power lines is also considered.

\begin{table}
    \centering
    \vspace{-0.6cm}
    \caption{Two environments considered in this study along with their characteristics}
    \resizebox{\linewidth}{!}{
    \begin{tabular}{l|c|c}
        \toprule
        Env name & IEEE 14 & NeurIPS 2020\\\hline
        Grid & \includegraphics[width=0.43\linewidth]{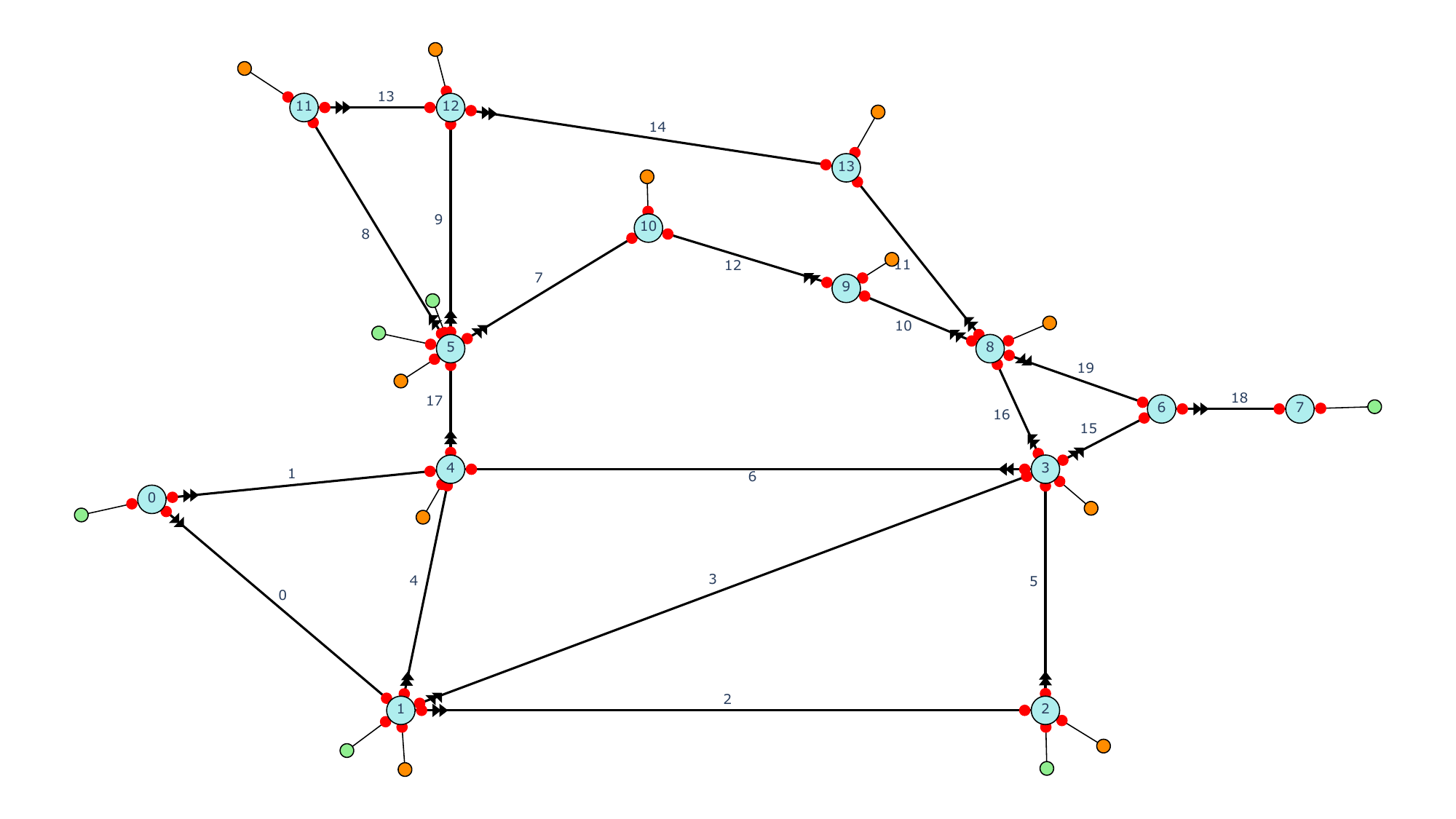} & \includegraphics[width=0.43\linewidth]{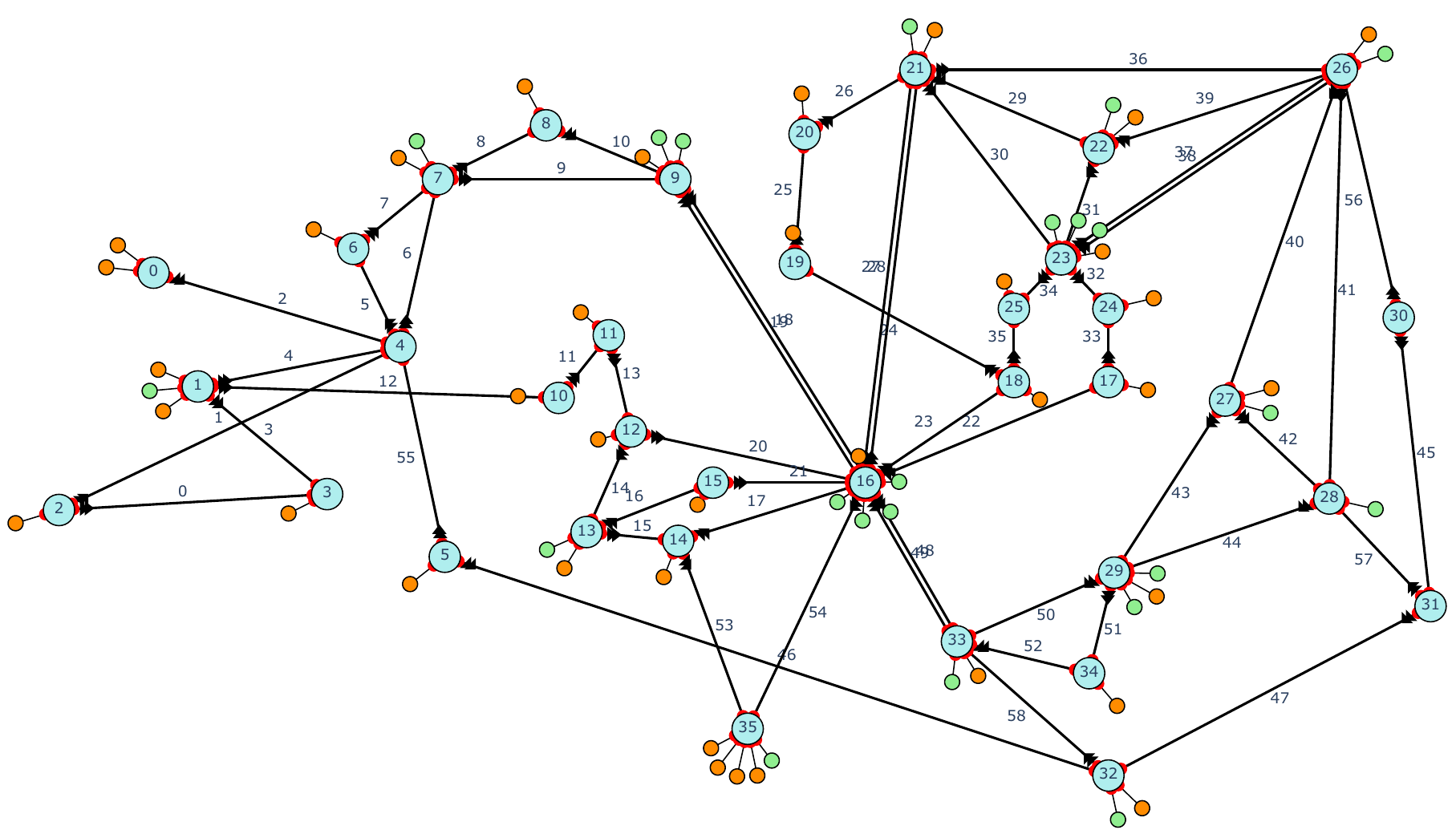}  \\\hline
         \# substations & 14 & 36 \\\hline
         \# lines & 20 & 58 \\\hline
         \# generators & 6 & 22 \\\hline
         \# loads & 11 & 37 \\
         \bottomrule
    \end{tabular}
    }
    \vspace{-0.5cm}
    \label{tab:envs}
\end{table}

A specific scenario is designed for the generation of datasets using a DC solver and \textit{lightsim2grid}\footnote{\url{https://github.com/Grid2op/lightsim2grid}} Python package. In this scenario, we consider a set of reference topology which are randomly applied to each sample (observation). The reference topology may include the grid configurations where the elements connected to each substation are randomly connected to one of two bus bars at each node. It could also include the original observation without any changes (with a predefined probability). On top of this reference topology, we consider a set of additional operations including power line disconnections and further bus bar reconfigurations. In total, four different datasets are generated representing the following characteristics:
\begin{itemize}
    \item[$\bullet$] \textit{Training dataset}: it includes 100,000 samples. In addition to the reference topologies, sampled randomly, it may include at most one line disconnection;
    \item[$\bullet$] \textit{Validation dataset}: it includes 10,000 samples and presents exactly the same distribution as the training set. This is used to analyze the over-fitting problem of ML-based approaches;
    \item[$\bullet$] \textit{Test dataset}: it includes 10,000 samples. In addition to the randomly sampled reference topologies, it includes one random power line disconnection for each observation;
    \item[$\bullet$] \textit{Out-of-distribution (OoD) dataset}: it includes 10,000 samples. It is considered as a secondary test data set to evaluate the generalization capability of the models on the observations represented by a slightly different distribution than the training set. In addition to reference topologies, it includes two simultaneous power lines disconnections which have never been considered in other datasets. 
\end{itemize}


\section{Study design}
\label{sec:study}
In this section, different neural network architectures along with their technical and theoretical details are described. We start with the simplest architecture without considering the physical constraints. It is followed by more complex architectures, where the physical constraints may be used as a regularization term or as the optimization objective. 
All the studies are accompanied by a schematic representation. In all these graphical representations, the first row presents the general steps, the corresponding visualization and architecture are shown in second row, and last row represents the details concerning the optimization objective (loss function).

\begin{figure}
    \centering
    \includegraphics[width=0.8\linewidth]{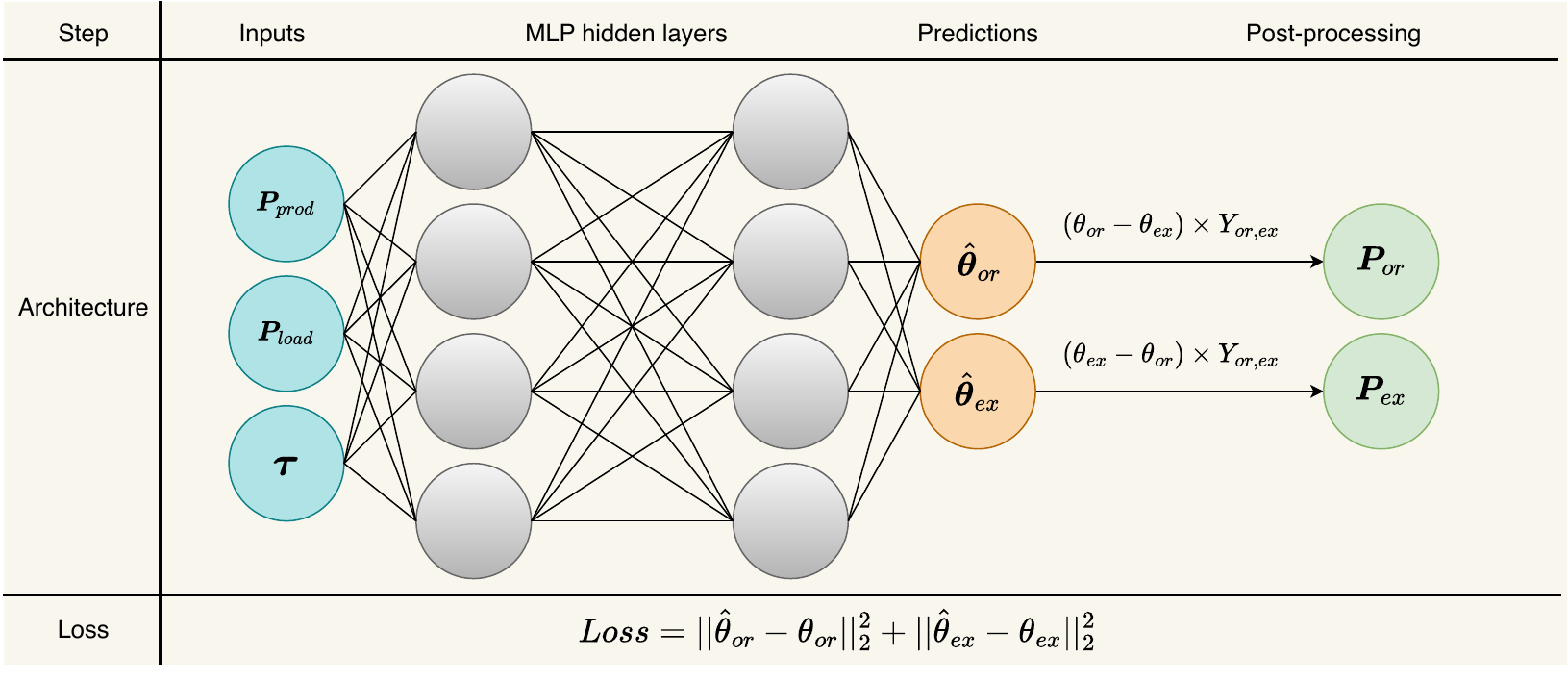}
    \caption{Multi-layer perceptron (MLP). The inputs $\boldsymbol{X}$ (injection + topology) in blue , the model predictions $\boldsymbol{\hat{y}} = (\hat{\theta}_{or}, \hat{\theta}_{ex})$ in orange (phasors) and post-processed outputs $(P_{or}, P_{ex}$) corresponding to active powers in green.}
    \label{fig:MLP}
\end{figure}

\paragraph{\textbf{Multi-layer perceptron (MLP)}} The first configuration consists in considering the power flow computation as a simple supervised regression problem. Figure \ref{fig:MLP} demonstrates the entire procedure for computing the power flow. As can be seen, the MLP takes the injections ($\boldsymbol{P}_{prod}$,  $\boldsymbol{P}_{load}$) alongside the topology vector ($\boldsymbol{\tau}$) \textit{i.e.}, the bus connectivity and line status as inputs, and predicts the phasors $\boldsymbol{\hat{\theta}} = MLP(\boldsymbol{P}_{prod}, \boldsymbol{P}_{load}, \boldsymbol{\tau})$ at the power lines. The active powers at the origin $\boldsymbol{P}_{or}$ and extremity $\boldsymbol{P}_{ex}$ sides of the power lines could be easily computed from the predicted phasors in a post-processing step. In this configuration, the loss function is simply the mean squared error between the predictions of the MLP and the ground-truth values.

\begin{figure*}[!h]
    \centering
    \includegraphics[width=1\columnwidth]{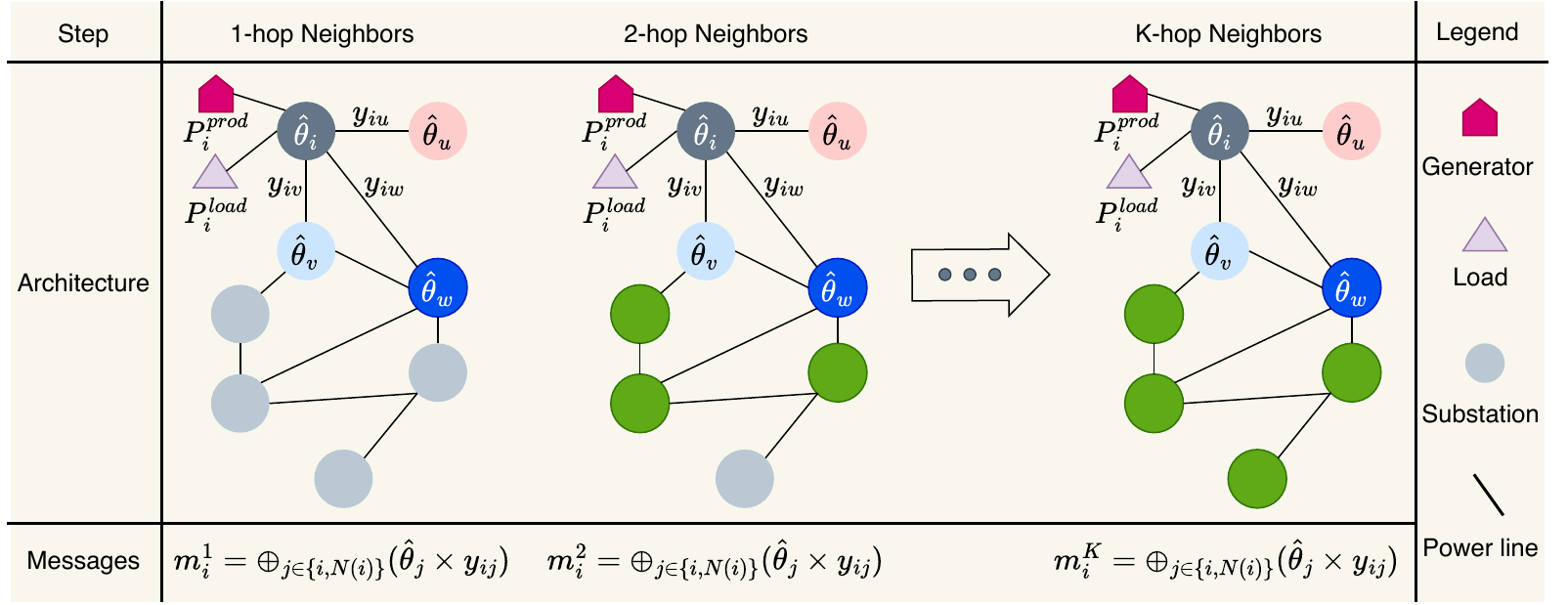}
    \caption{Message-Passing (MP) mechanism demonstration. The message includes the information from farther nodes by adding extra layers ($k$-hop neighborhood). The maximum number of effective layers is equal to $K$ which is equal to the number of nodes in the graph. In power grid context, each message represents the incoming power flow to a node $i$ from a neighboring node $j$ which is obtained by multiplying the phasor at node $j$ with the admittance value on the power line connecting two nodes ($\theta_j \times y_{ij}$).}
    \label{fig:message_passing}
\end{figure*}

\paragraph{\textbf{Message passing mechanism (MP)}} In this section, we provide the technical details concerning the message passing mechanism for power flow simulation, which enables the computation of physical constraint using graph theory. The classic message passing mechanism, as demonstrated in Figure \ref{fig:message_passing}, allows to propagate messages from the neighboring nodes. The propagation consists of three steps: message computation, aggregation, and update. By adding more layers (increasing the hops in the graph terminology), each message could include the information contained in farther nodes in the graph. This is shown by changing the nodes colors from light gray to green from the second hop neighbors.

\begin{figure*}[!h]
    \centering
    \includegraphics[width=0.9\columnwidth]{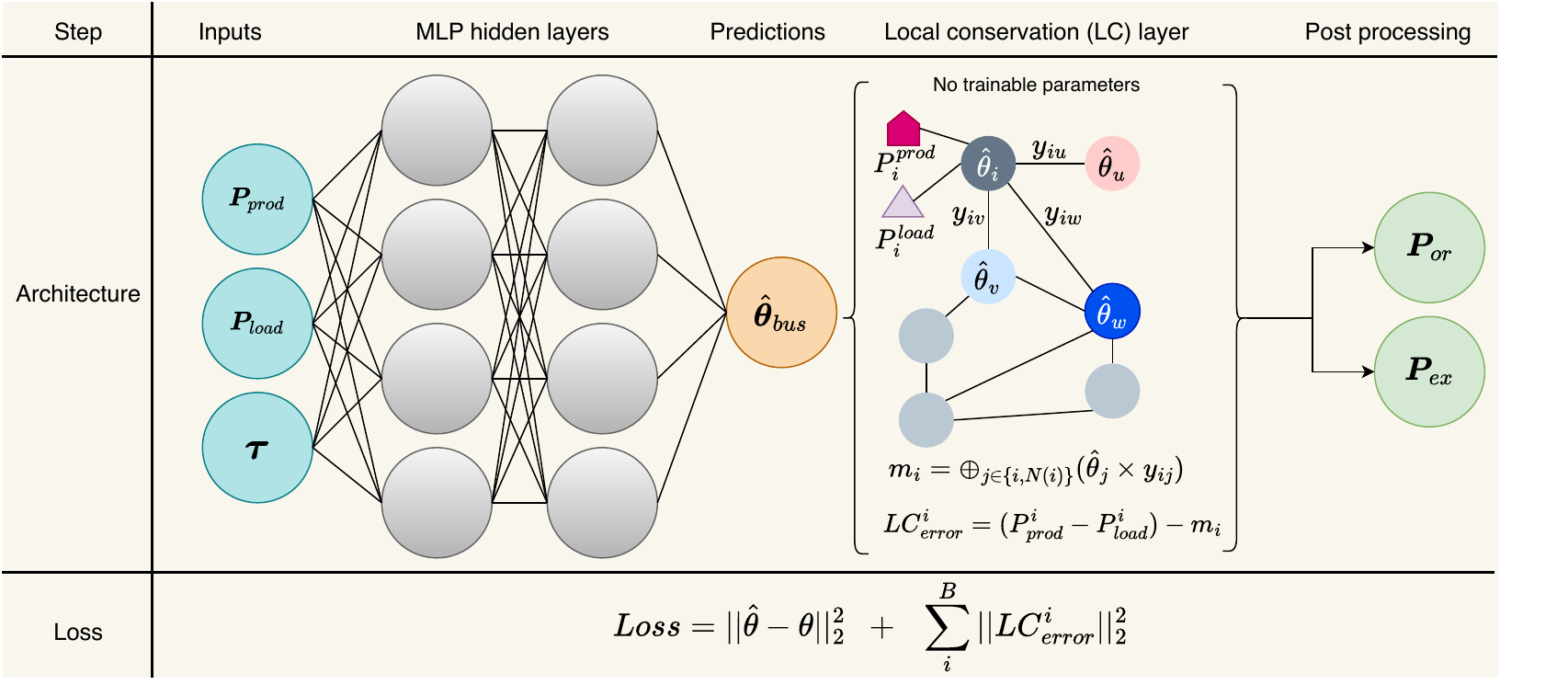}
    \caption{Regularized MLP. The inputs $\boldsymbol{X}$ (injections + topology) in blue, the model predictions $\boldsymbol{\hat{y}} = \{\hat{\theta}_{bus}\}_{bus\in(1,\ldots, B)}$ in orange (phasors) and post-processed outputs $(P_{or}, P_{ex}$) corresponding to active powers in green. A local conservation (message passing) layer is added after the predictions to compute physical constraint error which is considered as a regularization term in the loss function.}
    \label{fig:mlp_regularized}
\end{figure*}

However, this may not be the required behavior for power flow computation, as we want the messages to represent the underlying physics equation. As shown in the last row of Figure \ref{fig:message_passing}, each message is an aggregation of the information contained in neighboring nodes multiplied by the admittances on power lines. Alternatively, it represents the power flow that entered a specific substation. In the following, we exploit this message-passing mechanism in two different ways:
\begin{enumerate}
    \item \textit{To compute the local conservation law.} This is very helpful, as it allows to compute the physical constraint, which could be exploited by the auto-differentiation of neural networks. In the backward step, it enables the estimation of neural network parameters with respect to the gradients of local conservation law. (see Figure \ref{fig:mlp_regularized}); 
    \item \textit{To update the phasors using physics equation.} The same message passing mechanism could be used to compute the new estimations of the phasors. This is used by graph neural solvers, explained in next sections.
        
    
\end{enumerate}

\paragraph{\textbf{Regularized MLP (MLP Reg)}} The regularized MLP adds a regularization term into the loss function of a classical MLP and adopts a slightly different architecture, as can be seen in Figure \ref{fig:mlp_regularized}. It uses the same set of inputs as MLP, but to compute the local conservation error through the message-passing mechanism, it predicts the phasors at the bus level instead of the extremities of the power lines. The local conservation error $LC_{error}$ obtained using a single message passing layer is considered as a regularization term in the loss formulation and represents the physical constraint that the model output should satisfy. Once the phasors are predicted, the active powers at the power line extremities could be easily computed at the post-processing step.  

\begin{figure*}
    \centering
    \vspace{-0.5cm}
    \includegraphics[width=1\linewidth]{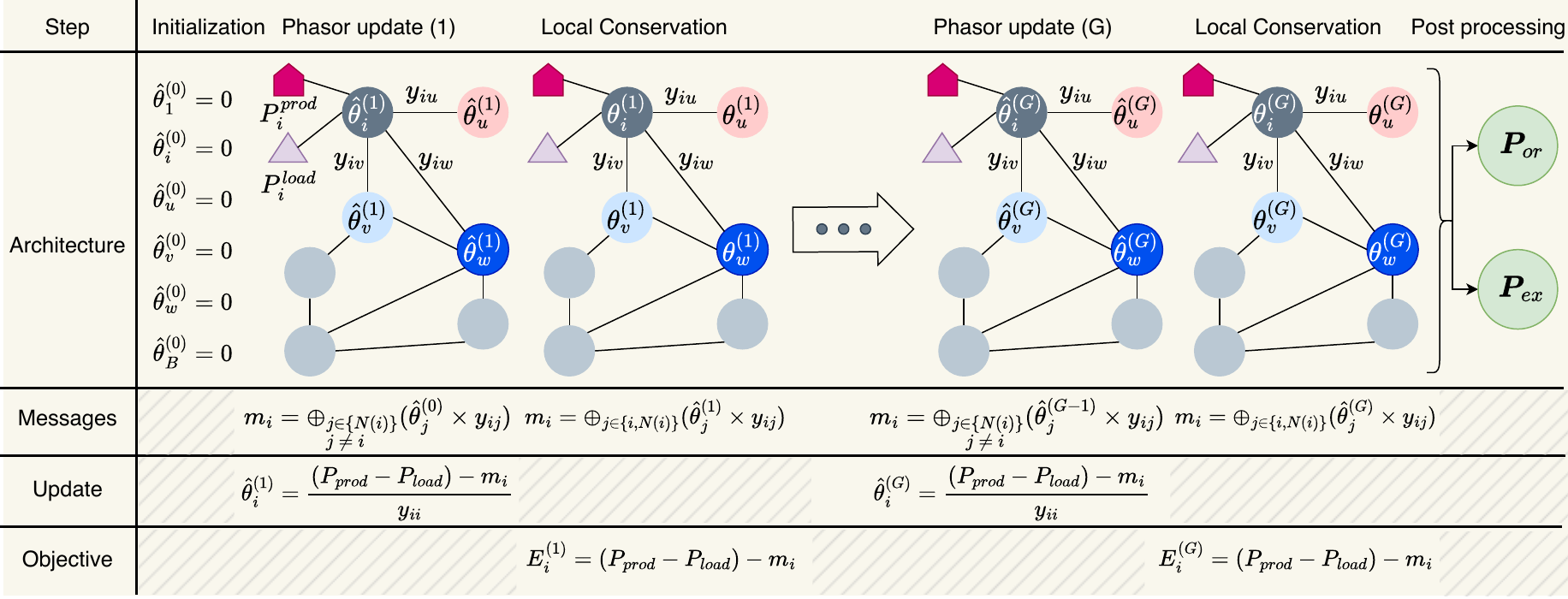}
    \caption{Message-passing as physics optimization with flat initialization of phasors (MP Opt). The architecture consists in interleaving two message-passing layers to compute the new $\theta$s (optimization) and local conservation error.}
    \label{fig:gnn_opt}
    \vspace{-0.5cm}
\end{figure*}

\paragraph{\textbf{Message-passing and physics constraint as optimization (MP Opt)}}
In this specific study, we consider the message-passing mechanism for both updating the phasor values and to compute the local conservation error. As can be seen in Figure \ref{fig:gnn_opt}, the phasors are initialized with zeros (flat initalization). Next, the two message-passing layers to update the phasors and to compute local conservation are interleaved. In the following, we explain the theoretical details of the update step.

The computation of new phasor values over message-passing layers is based on the local conservation law formulation, which for a given substation $i$ is given by:
\begin{equation}
\label{eq:lc_formula}
    p_i^{prod} - p_i^{load} = \sum_{\ell\in N(i)} p_i^\ell,
\end{equation}
where $p_i^\ell$ designates the power flow at a power line $\ell$ connected to a substation $i$. This is equivalent to:
\begin{equation}
\label{eq:lc_formula_ext}
    p_i^{prod} - p_i^{load} = \sum_{j\in \{i, N(i)\}} \theta_j \times y_{ij},
\end{equation}
where $\theta_j$ represents the phasor at a neighbor node $j$ and $y_{ij}$ is the admittance between two adjacent nodes $i$ and $j$ which is extracted from the admittance matrix $Y$. To compute the new phasor values at a node $i$, by considering $N(i)=\{u, v, w\}$ as its neighbors (see Figure \ref{fig:gnn_opt}), the Equation \ref{eq:lc_formula_ext} becomes:
\begin{equation}
\label{eq:lc_example}
    p_i^{prod} - p_i^{load} = (\theta_i \times y_{ii}) + {\color{mypink2} \underbrace{(\theta_u \times y_{iu})}_{\substack{\text{message from} \\ \text{node $u$}}}} + {\color{cornflowerblue} \underbrace{(\theta_v \times y_{iv})}_{\substack{\text{message from} \\ \text{node $v$}}}} + {\color{dollarbill}\underbrace{(\theta_w \times y_{iw})}_{\substack{\text{message from} \\ \text{node $w$}}}},
\end{equation}
where $y_{ii}$ is the admittance at the node $i$. As can be seen, the messages computed to update the phasor include only the information contained in the neighboring nodes, and not the graph node for which the update should be computed. This could be managed using self-loops in message-passing mechanism. Finally, the new value of our target, which is the phasor $\theta$ at node $i$ and for a layer $k$, is computed as follows:
\begin{equation}
    \theta_i^{(k)} = \frac{(p_i^{prod} - p_i^{load}) - [(\theta_u^{(k-1)} \times y_{iu}) + (\theta_v^{(k-1)} \times y_{iv}) + (\theta_w^{(k-1)} \times y_{iw})]}{y_{ii}}.
\end{equation}


Finally, the local conservation error could easily be calculated for each layer $k$ and at each node $i$ from Equation \eqref{eq:lc_formula_ext} as:
\begin{equation}
\label{eq:lc_formula_ext_ext}
    E_{i}^{(k)} = p_i^{prod} - p_i^{load} - \sum_{j\in \{i, N(i)\}} \theta_j \times y_{ij}.
\end{equation}

\begin{figure}[H]
    \centering
    \includegraphics[width=1\linewidth]{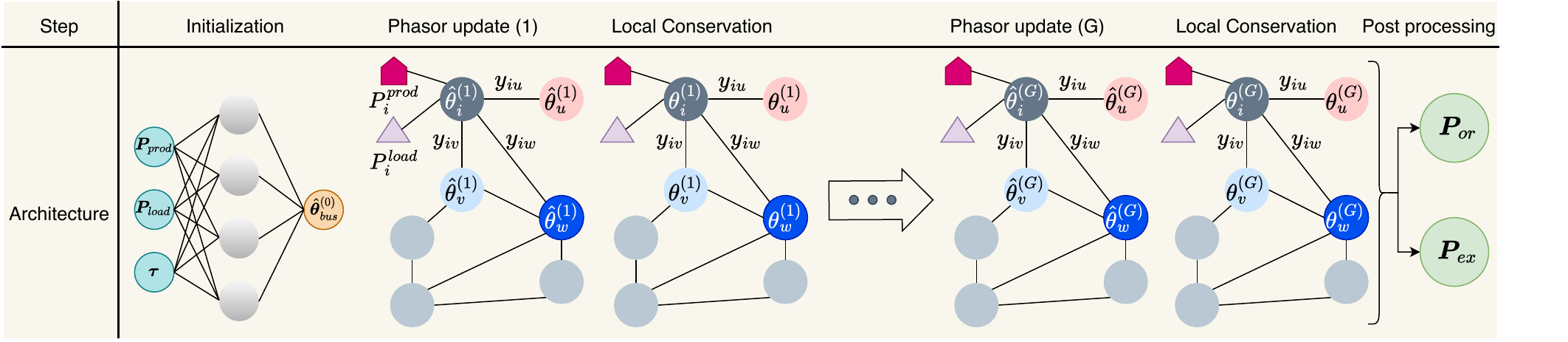}
    \caption{Physics-informed message-passing (PIMP) with warm initialization ($\hat{\theta}^{(0)}=MLP(P_{prod}, P_{load}, \tau; \omega)$). The messages, updates and objectives are exactly the same as the one shown in Figure \ref{fig:gnn_opt}}
    \label{fig:gnn_mlp}
\end{figure}

\paragraph{\textbf{Physics-informed message-passing with warm initialization (PIMP)}}
This last study consists in adding a learning paradigm into the optimization-based solution introduced in the previous study. As can be seen in Figure \ref{fig:gnn_mlp}, instead of flat initialization, the phasors are initialized using an MLP neural network (warm initialization). We expect that this initialization, by introducing learnable parameters, would reduce the required number of iterations (message-passing layers) for convergence. As such, back-propagation takes into account the graph operations when updating the MLP parameters, which would allow the better initialization of phasors by considering implicitly the physical constraint. In contrast to studies that exploit the GNNs \cite{lin2024powerflownet}, our approach relies on the resolution of physics equations. 

\section{Benchmark and evaluation}
\label{sec:benchmark}
This section introduces the experimental setups used by each technique during the evaluation, details the benchmark testbed and the evaluation criteria and presents the obtained results.

\subsection{Experimental settings}
The Direct Current (DC) solver from \textit{lightsim2grid} package is used for data generation and as the baseline for benchmarking. For more reliable results, all the experiments are executed five times and the results are reported using mean and standard deviation. The hyper-parameters of all the introduced techniques are fine-tuned using the Nevergrad library \cite{nevergrad}. 
Moreover, a dynamic learning rate reducing technique (i.e., reduce learning rate on plateau) is used, which exploits the validation loss for a better learning rate adjustment during the training of the models. The training of physics-informed models often starts with rapid improvements (as the model learns the rough shape of the solution). As it tries to satisfy stricter physics constraints, loss reduction slows down and avoids the overfitting and premature stagnation. The details concerning the hyper-parameters could be seen in the Github repository.

\subsection{Benchmark testbed and evaluation criteria}
The Learning Industrial Physical Simulation (LIPS) framework \cite{leyli2022lips} is used as the benchmark testbed and evaluation pipeline. It suggests to evaluate the performance of the models with respect to four categories of evaluation criteria:
\begin{itemize}
\renewcommand\labelitemi{$\bullet$}
    \item \textit{Machine learning related}: The precision of the models to predict the output variable (the phasors $\theta$) is computed using the mean absolute error (MAE) and mean absolute percentage error computed on the 10\% of highest values (MAPE90);
    \item \textit{Physics compliance}: To examine the physical compliance of different approaches, we propose to evaluate them with respect to a set of power grid physical laws considered as metrics. These evaluation criteria and the corresponding metrics are detailed in Table \ref{tab:Physical_laws}; 
    
\begin{minipage}{0.95\columnwidth}
    \begin{table}[H]
        \vspace{-0.8cm}
        \centering
        \caption{Physical constraints considered for the evaluation of \textit{physics compliance} category. The predictions should be conform, at best, to all of these constraints.}
        \renewcommand{\arraystretch}{1.3}
        \resizebox{\columnwidth}{!}{
        \begin{tabular}{cllm{5cm}}
            \toprule
            \textbf{ID} & \textbf{Type} & \textbf{Measure} & \textbf{Description} \\
            \hline
            \multicolumn{4}{c}{\textbf{Basic}} \\
            \hline
            \hline
            \vspace{.2cm} \underline{P1} & Losses positivity & $\frac{1}{L}\sum_\ell^L \mathds{1}_{(\hat{p}^\ell_{ex} + \hat{p}^\ell_{or} < 0.)}$ & Proportion of negative energy losses\\ 
            \hline
            \vspace{.2cm} \underline{P2} & Disconnected Line & $\frac{1}{L_{disc}}\sum_{\ell_{disc}}^{L_{disc}} \mathds{1}_{(\lvert\hat{x}^\ell_{ex}\rvert + \lvert\hat{x}^\ell_{or}\rvert > 0.)}$ & Proportion of non-null $a,p$ or $q$ values \\
            \hline
            \vspace{.2cm} \underline{P3} & Energy Losses & $\frac{\sum_{\ell=1}^L (\hat{p}^{(\ell)}_{ex} + \hat{p}^{(\ell)}_{or})}{Gen} \in [0.005,0.04]$ & Energy losses range consistency \\
            \hline
            \hline
            \multicolumn{4}{c}{\textbf{Uni-dimension Law}} \\
            \hline
            \vspace{.2cm} \underline{P4} & Global Conservation & $MAPE((Prod - Load) -  (\sum_{\ell=1}^L (\hat{p}^{\ell}_{ex} + \hat{p}^{\ell}_{or})))$ & Mean energy losses residual \\
            \hline
            \vspace{.2cm} \underline{P5} & Local Conservation & $MAPE((p^{prod}_k - p^{load}_k) - (\sum_{l \in neig(k)} \hat{p}^{\ell}_{k}))$ & Mean active power residual at nodes \\
            \bottomrule
        \end{tabular}}
        \label{tab:Physical_laws}
    \end{table}
\end{minipage}
\vspace{0.3cm}

    \item \textit{Out-of-distribution generalization}: To evaluate the generalization capability of the models, the same machine learning metrics are computed using the OoD dataset. This dataset includes a slightly different distribution (two simultaneous disconnected power lines) compared to training and test datasets (only one power line disconnection);
    \item \textit{Industrial readiness}: To evaluate the industrialization capability of the models, we have considered the acceleration or speed-up over the physical solver (DC solver). To consider the scaling capability of different studies, we have also considered two power grids of different sizes, namely IEEE 14 and NeurIPS small including 36 substations (see Table \ref{tab:envs}).
\end{itemize}

\subsection{Results}
Table \ref{tab:benchmark_table} summarizes the benchmark results for the various studies introduced in the previous section.  
These studies are compared through the four previously mentioned categories of evaluation criteria. 
The performances are computed between the predicted phasors ($\hat{\theta}$) and ground-truth values (DC-solver). For the physics compliance category, we only consider the local conservation error in terms of the violation percentage ($\underline{P5}$ in Table \ref{tab:Physical_laws}), as all the methods satisfy the remaining considered physical constraints. The best performances for training-based approaches (excluding the MP Opt) are highlighted in bold.

\begin{table}
    \centering
    \vspace{-0.5cm}
    \caption{Benchmark table. The results are presented using mean $\pm$ standard deviation (over 5 runs). The compared methods are multi-layer perceptrons (MLP), MLP with physical constraint as regularization (MLP Reg), message-passing  and physics constraint as optimization (MP Opt) and physics-informed message-passing with warm initialization (PIMP). The output variables are phasors at each node of the graph ($\theta$). The considered criteria are Mean Absolute Error (MAE) and Mean Absolute Percentage Error (MAPE) computed on 10\% of lines with highest values (MAPE90), Inference time represented by the speed-up with respect to physic solver (Inf. speed-up), Out-of-distribution Generalisation (OOD Gen.), and physics compliance category (Physics comp.) computing (P5) local conservation laws on both test and OOD datasets.}
    \resizebox{\linewidth}{!}{
    \begin{tabular}{clccccccccccc}
    \toprule
         & & & \multicolumn{10}{c}{\textbf{Evaluation Criteria Categories}}\\
         & & & \multicolumn{2}{c}{\textbf{ML-related}} & & \textbf{Readiness} & & \multicolumn{2}{c}{\textbf{OOD Gen.}} & & \multicolumn{2}{c}{\textbf{Physics Comp.}} \\\cline{4-5} \cline{7-7} \cline{9-10} \cline{12-13}
         & \textbf{Methods} & \textbf{Variable} & MAE & MAPE90  && Inf. speed-up && MAE & MAPE90 && \ \ \ \ Test\ \ & \ \ OOD\\ \cline{1-13} 
         \multicolumn{1}{c|}{\multirow{8}{*}{\rotatebox{90}{\texttt{IEEE 14}}}} & \multicolumn{1}{c|}{}\\

         \multicolumn{1}{c|}{} & \multicolumn{1}{c|}{MLP} & $\theta$ &  1e-2$\pm$1e-3 & 2e-3$\pm$2e-4  && \textbf{4} && 4e-1$\pm$1e-2 & 1e-1$\pm$2e-3 && 40\%$\pm$1\% & 47\%$\pm$.5\%\\ \cline{3-13} \multicolumn{1}{c|}{} & \multicolumn{1}{c|}{} & \multicolumn{1}{c}{}\\
         \multicolumn{1}{c|}{} & \multicolumn{1}{c|}{MLP Reg} & $\theta$ &  6e-2$\pm$2e-2 & 1e-2$\pm$5e-3 && \textbf{4} && 6e-1$\pm$4e-2 & 1e-1$\pm$9e-3 && 35\%$\pm$1\% & 42\%$\pm$1\%\\ \cline{3-13} \multicolumn{1}{c|}{} & \multicolumn{1}{c|}{} & \multicolumn{1}{c}{}\\
         \multicolumn{1}{c|}{} & \multicolumn{1}{c|}{MP Opt} & $\theta$ & 8e-4 $\pm$ 7e-9 & 4e-3 $\pm$ 7e-9 && 1 && 1e-2$\pm$3e-9 & 4e-3$\pm$8e-9 && 0\% & 0\%\\ \cline{3-13} \multicolumn{1}{c|}{} & \multicolumn{1}{c|}{} & \multicolumn{1}{c}{}\\
         \multicolumn{1}{c|}{} & \multicolumn{1}{c|}{PIMP} & $\theta$ &  \textbf{2e-3$\pm$2e-5} & \textbf{7e-4$\pm$6e-6} && 2 && \textbf{2e-2$\pm$1e-4} & \textbf{8e-3$\pm$5e-5} && \textbf{1\%$\pm$.1\%} & \textbf{13\%$\pm$.7\%} \\ \cline{1-13}
         \multicolumn{1}{c|}{\multirow{8}{*}{\rotatebox{90}{\texttt{NeurIPS (36 nodes)}}}} & \multicolumn{1}{c|}{}\\

         \multicolumn{1}{c|}{} & \multicolumn{1}{c|}{MLP} & $\theta$ &  1e-1$\pm$2e-3 & 5e-2$\pm$1e-3  && \textbf{8} && 2e-1$\pm$2e-3 & 1e-1$\pm$1e-3 && 50\%$\pm$ 1\% & 50\%$\pm$1\%\\ \cline{3-13} \multicolumn{1}{c|}{} & \multicolumn{1}{c|}{} & \multicolumn{1}{c}{}\\
         \multicolumn{1}{c|}{} & \multicolumn{1}{c|}{MLP Reg} & $\theta$ &  7e-1$\pm$2e-1 & 4e-1$\pm$8e-2 && \textbf{8} && 9e-1$\pm$2e-2 & 5e-1$\pm$8e-2 && 45\%$\pm$1\% & 47\%$\pm$1\%\\ \cline{3-13} \multicolumn{1}{c|}{} & \multicolumn{1}{c|}{} & \multicolumn{1}{c}{}\\
         \multicolumn{1}{c|}{} & \multicolumn{1}{c|}{MP Opt} & $\theta$ &  4e-3$\pm$1e-9 & 2e-3$\pm$6e-9 && 1 && 1e-2$\pm$1e-9 & 4e-3$\pm$4e-9 && 0\% & 0\% \\ \cline{3-13} \multicolumn{1}{c|}{} & \multicolumn{1}{c|}{} & \multicolumn{1}{c}{}\\
         \multicolumn{1}{c|}{} & \multicolumn{1}{c|}{PIMP} & $\theta$ &  \textbf{4e-2$\pm$3e-5} & \textbf{2e-2$\pm$2e-3} && 4 && \textbf{6e-2$\pm$1e-3} & \textbf{3e-2$\pm$4e-3} && \textbf{20\%$\pm$.5\%} & \textbf{26\%$\pm$.5\%} \\
         \bottomrule
    \end{tabular}}
    \vspace{-0.5cm}
    \label{tab:benchmark_table}
\end{table}

\paragraph{\textbf{Analysis of IEEE14 environment}}
We can observe that the best results are achieved using the \textit{MP Opt}, which updates the target values based on physics equations and yields a very low standard deviation.
It may seem to be the normal behavior, as this approach does not include any stochastic initialization or operation during the optimization. It requires 100 iterations (message-passing layers) to converge to the optimal solution, where the local conservation error is very close to zero. 
One of the objectives of considering this study was to analyze whether the addition of some learnable parameters for the initialization phasors $\theta^{(0)}$ may allow reducing the number of required message-passing layers.

To conduct this experimentation, we have reduced the number of message-passing layers by half (60 layers) and added an MLP with four hidden layers for the initialization of phasors (\textit{PIMP}). It shows significantly better performances than MLP-based approaches and is comparable to that of the  pure optimization-based approach. The main difference could be seen with respect to the physics compliance criteria. Despite the notable improvement obtained over MLP-based approaches (reducing the local conservation error by more than 30\%), the compliance remains lower than MP Opt for the OoD dataset (13\% of law violation). Adding more message-passing layers allows to improve further the physics compliance, which indeed is directly related to the model complexity. 

The two MLP-based approaches obtain very close performances with respect to ML-related criteria for both test and OoD datasets. They show also the best speed-ups behavior with respect to the DC physical solver. The regularization through the penalization of the physics term allows to reduce slightly the local conservation error for \textit{MLP Reg}. During the experimentation, we have observed that this approach requires more iterations (epochs) for the better convergence than the MLP without any regularization. It should also be mentioned that the DC solver implemented in \textit{lightsim2grid} is very fast and the optimization could be computed in parallel for the considered scenario (a scenario known also as security analysis). This makes acceleration improvement quite challenging.

\paragraph{\textbf{Analysis of NeurIPS environment (36 nodes)}}
A similar pattern could also be observed for the larger grid including 36 substations. For this more complex power grid, the \textit{MP Opt} method requires substantially more iterations to converge (350 iterations or message-passing layers). The performance slightly degrades compared to the smaller grid across all evaluation criteria, with the exception of the speed-up. This highlights the role that hybrid models can play when applied to more realistic, large-scale grids. For the sake of this experimentation, the number of message passing layers (300 message passing layers) required by \textit{MP Opt} to fully satisfy the physical constraints is halved for \textit{PIMP}. We can observe clearly much better compliance with local conservation laws compared to MLP-based approaches. At the same time, an increasing trend in physical constraint violation could be observed which is correlated to network complexity, and may necessitate a more rigorous enforcement of such physical constraints.

\begin{figure*}[!b]
 \begin{subfigure}{0.49\textwidth}
     \includegraphics[width=\textwidth]{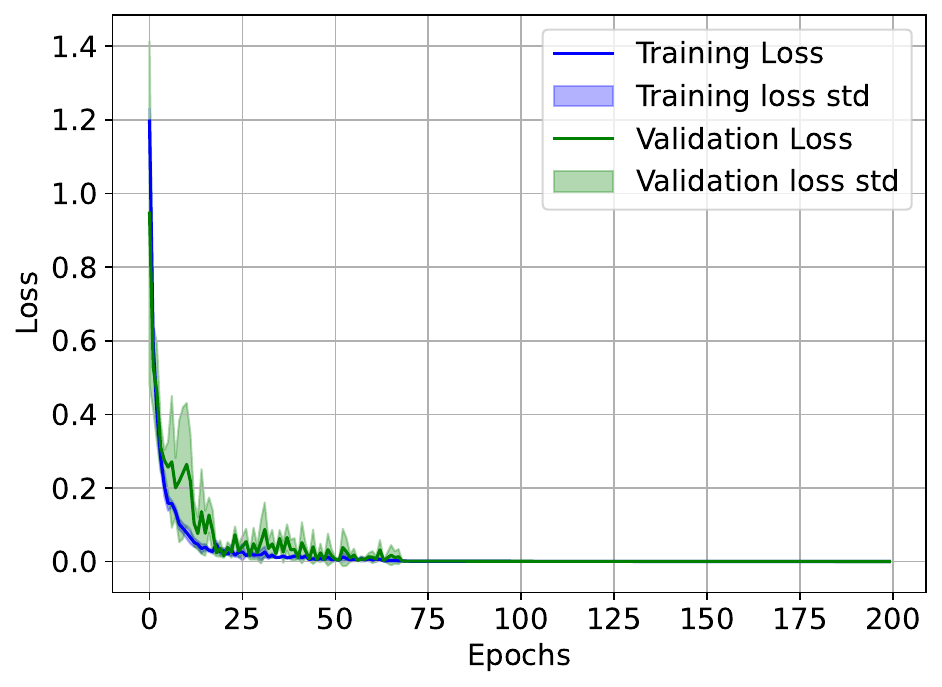}
     \caption{MLP}
     \label{fig:MLP_convergence}
 \end{subfigure}
 \hfill
 \begin{subfigure}{0.49\textwidth}
     \includegraphics[width=\textwidth]{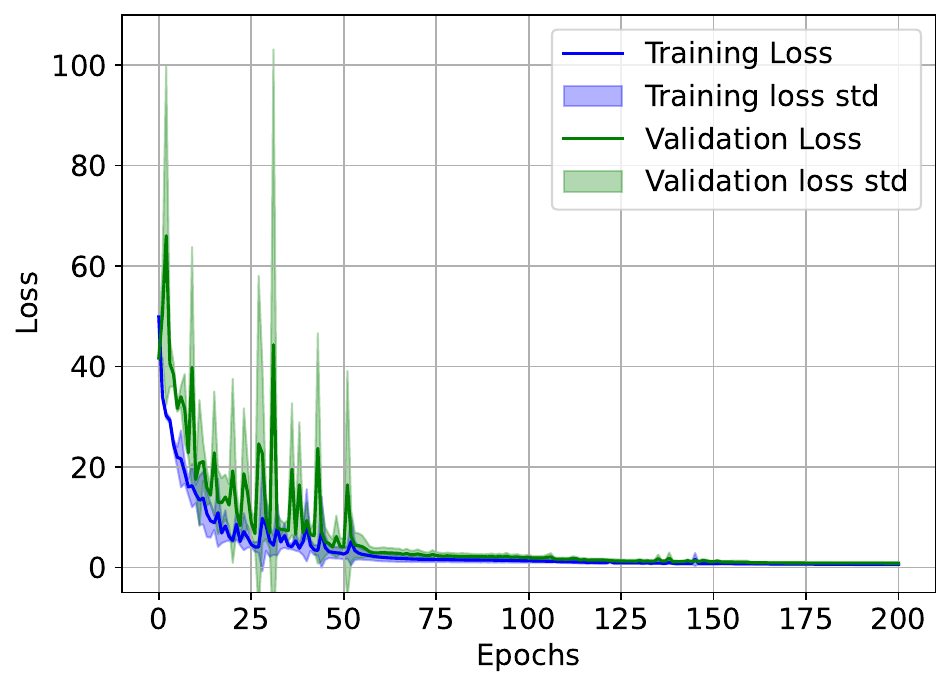}
     \caption{MLP regularized}
     \label{fig:MLP_reg_convergence}
 \end{subfigure}
 
 \medskip
 \begin{subfigure}{0.49\textwidth}
     \includegraphics[width=\textwidth]{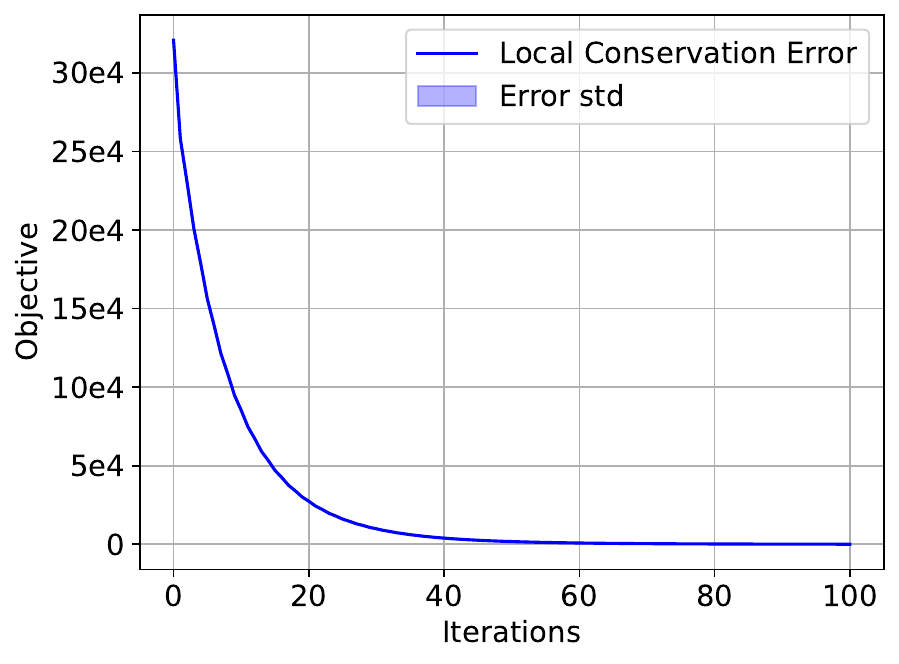}
     \caption{MP Opt}
     \label{fig:GNN_opt_convergence}
 \end{subfigure}
 \hfill
 \begin{subfigure}{0.49\textwidth}
     \includegraphics[width=\textwidth]{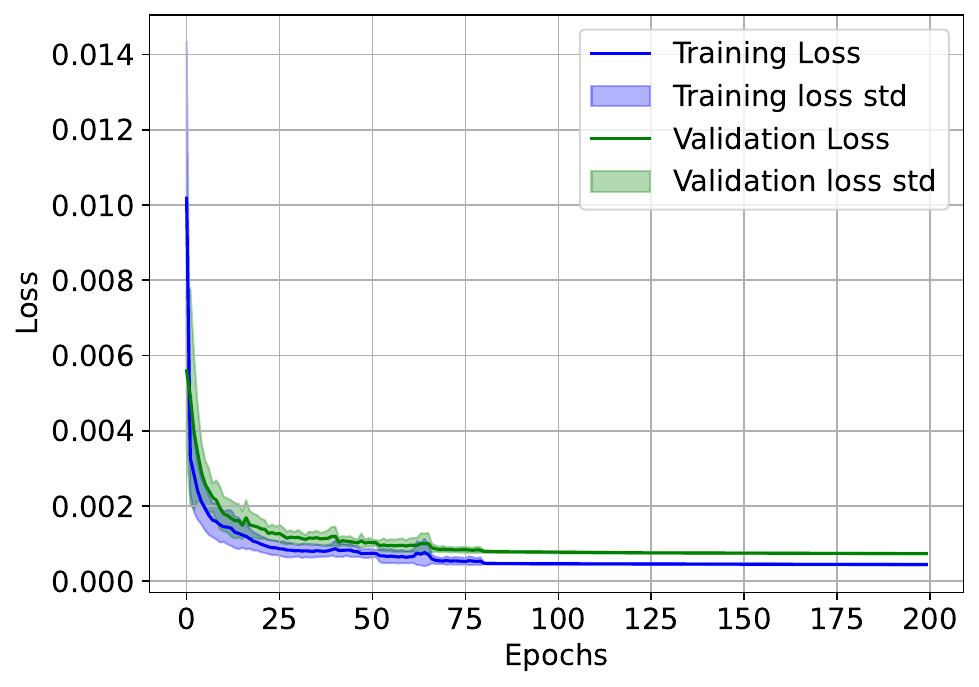}
     \caption{PIMP with MLP initialization}
     \label{fig:GNN_init_convergence}
 \end{subfigure}

 \caption{Convergence behavior of different approaches for IEEE 14 environment}
 \label{fig:convergence}

\end{figure*}

\paragraph{\textbf{Convergence analysis}} The convergence behavior of these approaches are compared in Figure \ref{fig:convergence} for IEEE14 environment. The convergence is represented using mean losses (solid lines), and the highlighted area around the mean represents the standard deviation over 5 runs. For the approaches requiring training, the validation loss is also shown in green in addition to the training loss which is shown in blue. It can be observed that all the approaches converge to a local optimum with slightly different convergence behaviors. The MLP-based approach (see Figure \ref{fig:MLP_convergence}) which does not consider the physical constraint converges enough fast with low variation of the loss over the data. However, the regularized MLP which considers the physical constraint as penalization of the loss, shows some variations in the beginning of the training due to the high values of local conservation error, and stabilizes when the predictions approach the ground-truth values (see Figure \ref{fig:MLP_reg_convergence}). The \textit{MP Opt} approach does not requiring training and optimizes the local conservation error directly through the message passing layers and updating the phasors. Finally, the \textit{PIMP} with warm initialization (see Figure \ref{fig:GNN_init_convergence}) presents more stabilized training curve than those obtained by MLP, and converges faster than the MP optimization-based approach.

\begin{figure}
 \begin{subfigure}{0.49\textwidth}
     \includegraphics[width=\textwidth]{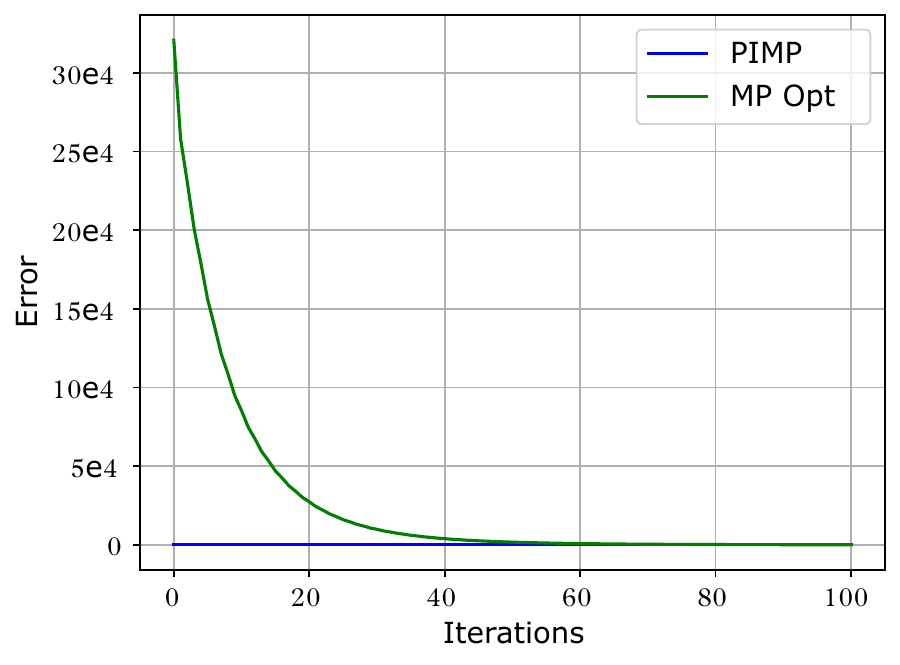}
     \caption{Natural scale}
     \label{fig:GNN_comaprison_natural}
 \end{subfigure}
 \hfill
 \begin{subfigure}{0.49\textwidth}
     \includegraphics[width=\textwidth]{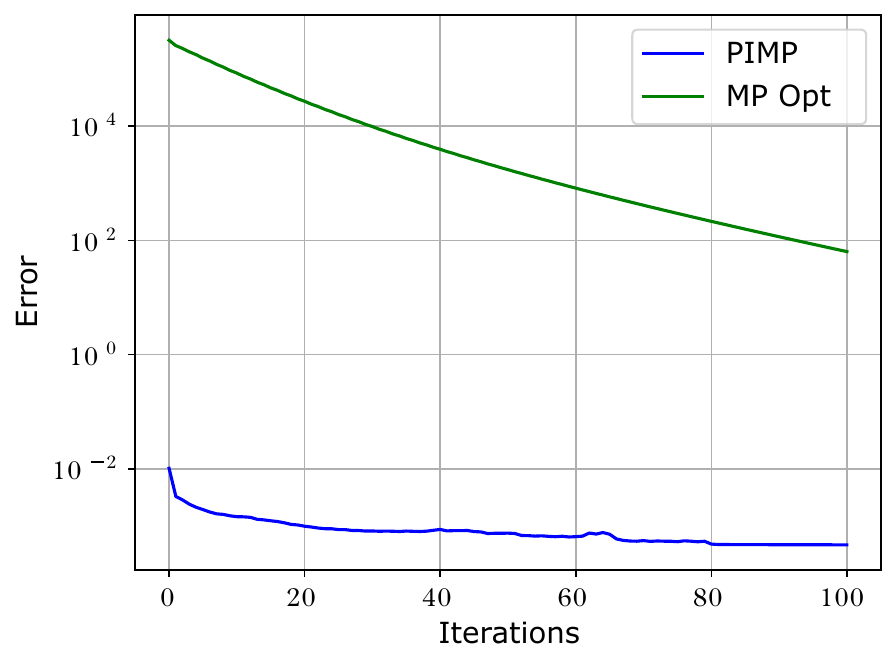}
     \caption{Log scale}
     \label{fig:GNN_comparison_log}
 \end{subfigure}

 \caption{Comparison of the training curve convergence between Message-Passing with optimization (\textit{MP Opt}) and Message-Passing with MLP initialization (\textit{PIMP}).}
 \label{fig:convergence_comparison}

\end{figure}

To analyze the difference between two Message-passing based approaches in greater details, Figure \ref{fig:convergence_comparison} compares their convergence using the same graphic (with both natural and log scale of local conservation errors on y-axis). For consistency, we have considered the same number of iterations for this analysis. As can be seen, the initialization of phasors using MLP enables the faster convergence of the approach (PIMP), despite the reduced number of message-passing layers. 

\begin{figure}
 \begin{subfigure}{0.47\textwidth}
     \includegraphics[width=\textwidth]{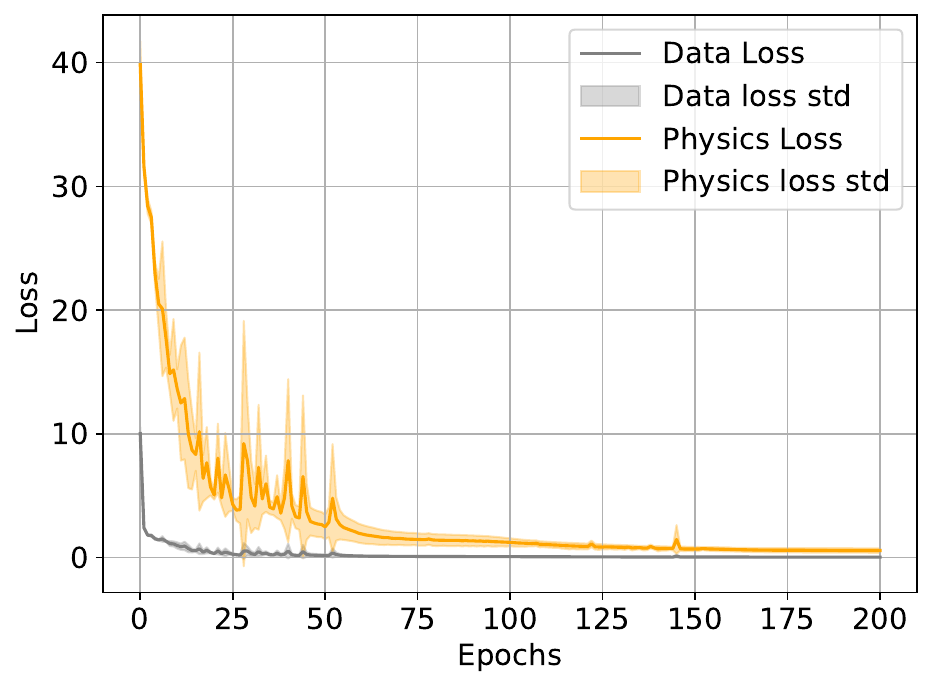}
     \caption{MLP regularized}
     \label{fig:mlp_reg_loss}
 \end{subfigure}
 \hfill
 \begin{subfigure}{0.49\textwidth}
     \includegraphics[width=\textwidth]{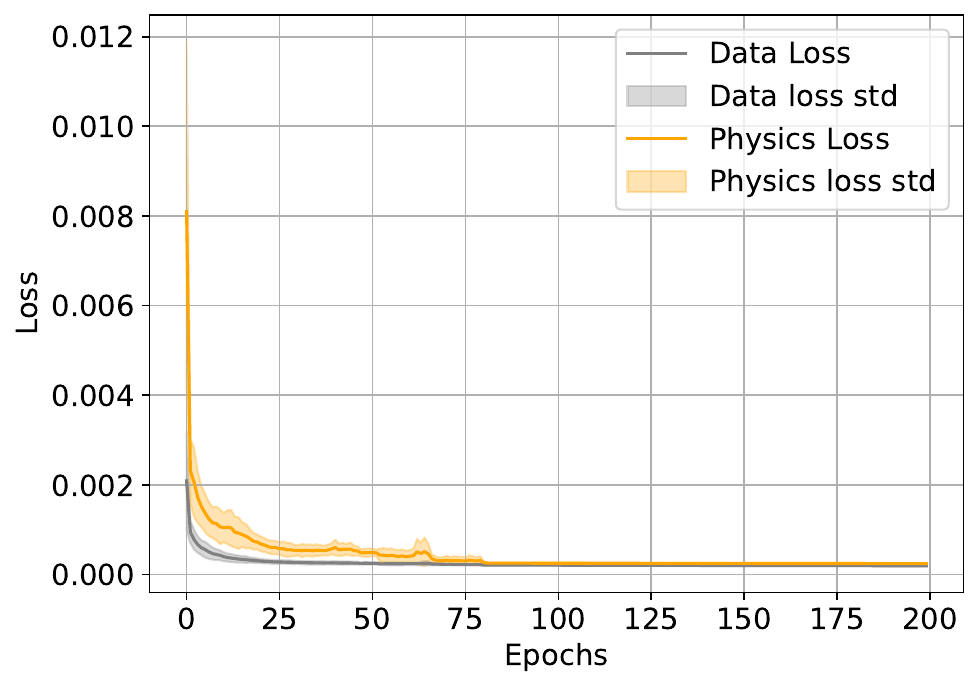}
     \caption{PIMP initialized using MLP}
     \label{fig:gnn_loss}
 \end{subfigure}

 \caption{Convergence behavior of physics loss vs. data loss}
 \label{fig:convergence_physics_data}
 \vspace{-0.9cm}
\end{figure}

Finally, to better understand the impact of data and physics terms during the training of \textit{MLP Reg} and \textit{PIMP} initialized using MLP, Figure \ref{fig:convergence_physics_data} demonstrates their behavior in separate graphics. As can be seen in Figure \ref{fig:mlp_reg_loss}, the physics loss (orange curve) for \textit{MLP Reg} shows high variability at the beginning of the training which stabilizes over time. However, it never reaches the optimal solution. On the other hand, \textit{PIMP} shows more stabilized physics loss behavior from the beginning of the training and overlaps the data loss early. It could be explained by the fact that the message-passing based approaches takes advantage of the data structure (grid topology) and allows to be consistent to physical context by design.

\section{Conclusion and perspectives}
\label{sec:conclusion}
This paper proposed a study design to benchmark and investigate the inner workings of physics-informed neural networks. A unified benchmarking framework was employed to enable the comprehensive evaluation of various methods. The experimental results indicate that pure machine learning-based techniques can achieve high accuracy; however, their performance often fall short in meeting real-world requirements, particularly in terms of physics compliance and out-of-distribution generalization. In contrast, approaches that incorporate physics knowledge -- either as a regularization term in the loss function or as part of the optimization objective -- demonstrated superior performance across most evaluation criteria.

Furthermore, the experiments revealed that physics-informed approaches exhibit strong potential in low-data regimes. These methods outperformed classical approaches when the number of training samples was reduced. All studies presented in this paper are fully reproducible, with implementations available in the GitHub repository, and are evaluated using a standardized evaluation pipeline, thereby supporting further research in this domain.

For didactic purposes, this study focused on modeling power flow using a DC solver. As a direction for future work, we plan to extend this framework to the more complex AC power flow, which additionally requires balancing voltage magnitudes. In the current work, equal importance was assigned to the data and physics components of the loss function. However, several techniques have been proposed in the literature to improve convergence and enhance physics compliance. For instance, gradually increasing the weight of the physics-based loss term has been shown to be effective \cite{krishnapriyan2021characterizing, hanna2024improved}. 

The application on real-world data may often encounter inconsistencies or other data issues. We plan to mitigate them by incorporating the mechanisms for handling missing or corrupted data through, for example, robust loss functions or anomaly detection, improving the solver stability and results reliability. We also observed that increasing grid complexity leads message-passing-based methods to require more iterations (or layers) to converge. Consequently, their computational complexity may increase linearly with the grid complexity. Future studies will focus on incorporating learnable parameters into the message-passing mechanism, moving toward the use of graph neural networks. This approach is expected to reduce the number of required iterations. 

\section*{Acknowledgments}
The research leading to this work is part of the AI4REALNET (\textit{AI for REAL-world NETwork operation}) project, which received funding from European Union’s Horizon Europe Research and Innovation programme under the Grant Agreement No 101119527, and from the Swiss State Secretariat for Education, Research and Innovation (SERI). This project is funded by the European Union and SERI. Views and opinions expressed are however those of the author(s) only and do not necessarily reflect those of the European Union and SERI. Neither the European Union nor the granting authority can be held responsible for them.
%
%
%
\bibliographystyle{splncs04}
\bibliography{references}

\begin{thebibliography}{10}
\providecommand{\url}[1]{\texttt{#1}}
\providecommand{\urlprefix}{URL }
\providecommand{\doi}[1]{https://doi.org/#1}

\bibitem{babatunde2020power}
Babatunde, O.M., Munda, J.L., Hamam, Y.: Power system flexibility: A review. Energy Reports  \textbf{6},  101--106 (2020)

\bibitem{baker2019learning}
Baker, K.: Learning warm-start points for ac optimal power flow. In: 2019 IEEE 29th International Workshop on Machine Learning for Signal Processing (MLSP). pp.~1--6. IEEE (2019)

\bibitem{battaglia2018relational}
Battaglia, P.W., Hamrick, J.B., Bapst, V., Sanchez-Gonzalez, A., Zambaldi, V., Malinowski, M., Tacchetti, A., Raposo, D., Santoro, A., Faulkner, R., et~al.: Relational inductive biases, deep learning, and graph networks. arXiv preprint arXiv:1806.01261  (2018)

\bibitem{carleo2019machine}
Carleo, G., Cirac, I., Cranmer, K., Daudet, L., Schuld, M., Tishby, N., Vogt-Maranto, L., Zdeborov{\'a}, L.: Machine learning and the physical sciences. Reviews of Modern Physics  \textbf{91}(4),  045002 (2019)

\bibitem{donnot2019leap}
Donnot, B., Donon, B., Guyon, I., Liu, Z., Marot, A., Panciatici, P., Schoenauer, M.: Leap nets for power grid perturbations. arXiv preprint arXiv:1908.08314  (2019)

\bibitem{donon2020neural}
Donon, B., Cl{\'e}ment, R., Donnot, B., Marot, A., Guyon, I., Schoenauer, M.: Neural networks for power flow: Graph neural solver. Electric Power Systems Research  \textbf{189},  106547 (2020)

\bibitem{hanna2024improved}
Hanna, J.M., Vignon-Clementel, I.E., Talbot, H.: Improved physics-informed neural networks loss function regularization with a variance-based term. Available at SSRN 5269341  (2024)

\bibitem{hasan2020survey}
Hasan, F., Kargarian, A., Mohammadi, A.: A survey on applications of machine learning for optimal power flow. In: 2020 IEEE Texas Power and Energy Conference (TPEC). pp.~1--6. IEEE (2020)

\bibitem{kasim2021building}
Kasim, M., Watson-Parris, D., Deaconu, L., Oliver, S., Hatfield, P., Froula, D., Gregori, G., Jarvis, M., Khatiwala, S., Korenaga, J., et~al.: Building high accuracy emulators for scientific simulations with deep neural architecture search. Machine Learning: Science and Technology  \textbf{3}(1),  015013 (2021)

\bibitem{krishnapriyan2021characterizing}
Krishnapriyan, A., Gholami, A., Zhe, S., Kirby, R., Mahoney, M.W.: Characterizing possible failure modes in physics-informed neural networks. Advances in neural information processing systems  \textbf{34},  26548--26560 (2021)

\bibitem{lam2023learning}
Lam, R., Sanchez-Gonzalez, A., Willson, M., Wirnsberger, P., Fortunato, M., Alet, F., Ravuri, S., Ewalds, T., Eaton-Rosen, Z., Hu, W., et~al.: Learning skillful medium-range global weather forecasting. Science  \textbf{382}(6677),  1416--1421 (2023)

\bibitem{lei2020data}
Lei, X., Yang, Z., Yu, J., Zhao, J., Gao, Q., Yu, H.: Data-driven optimal power flow: A physics-informed machine learning approach. IEEE Transactions on Power Systems  \textbf{36}(1),  346--354 (2020)

\bibitem{leyli2022lips}
LEYLI~ABADI, M., Marot, A., Picault, J., Danan, D., Yagoubi, M., Donnot, B., Attoui, S., Dimitrov, P., Farjallah, A., Etienam, C.: Lips-learning industrial physical simulation benchmark suite. Advances in Neural Information Processing Systems  \textbf{35},  28095--28109 (2022)

\bibitem{lin2024powerflownet}
Lin, N., Orfanoudakis, S., Cardenas, N.O., Giraldo, J.S., Vergara, P.P.: Powerflownet: Power flow approximation using message passing graph neural networks. International Journal of Electrical Power \& Energy Systems  \textbf{160},  110112 (2024)

\bibitem{lopez2023power}
Lopez-Garcia, T.B., Dom{\'\i}nguez-Navarro, J.A.: Power flow analysis via typed graph neural networks. Engineering Applications of Artificial Intelligence  \textbf{117},  105567 (2023)

\bibitem{raissi2017physics}
Raissi, M., Perdikaris, P., Karniadakis, G.E.: Physics informed deep learning (part i): Data-driven solutions of nonlinear partial differential equations. arXiv preprint arXiv:1711.10561  (2017)

\bibitem{nevergrad}
Rapin, J., Teytaud, O.: {Nevergrad - A gradient-free optimization platform}. \url{https://GitHub.com/FacebookResearch/Nevergrad} (2018)

\bibitem{saadat1999power}
Saadat, H., et~al.: Power system analysis, vol.~2. McGraw-hill (1999)

\bibitem{stott1974review}
Stott, B.: Review of load-flow calculation methods. Proceedings of the IEEE  \textbf{62}(7),  916--929 (1974)

\bibitem{tompson2016}
Tompson, J., Schlachter, K., Sprechmann, P., Perlin, K.: Accelerating eulerian fluid simulation with convolutional networks. ArXiv: 1607.03597  (2016)

\bibitem{vlach1983computer}
Vlach, J., Singhal, K.: Computer methods for circuit analysis and design. Springer Science \& Business Media (1983)

\bibitem{xavier2021learning}
Xavier, {\'A}.S., Qiu, F., Ahmed, S.: Learning to solve large-scale security-constrained unit commitment problems. INFORMS Journal on Computing  \textbf{33}(2),  739--756 (2021)

\end{thebibliography}

\end{document}